\documentclass[10pt,journal,compsoc]{IEEEtran}
\ifCLASSOPTIONcompsoc
  \usepackage[nocompress]{cite}
\else
  \usepackage{cite}
\fi
\ifCLASSINFOpdf
\else
\fi

\usepackage{xcolor}
\usepackage{subfigure}
\usepackage{xparse}
\usepackage{relsize}
\usepackage{hyperref}
\usepackage{bbm}
\usepackage{stmaryrd}
\usepackage{amsmath}
\usepackage{booktabs} 

\usepackage{xspace}
\usepackage[backgroundcolor=white]{todonotes}
\usepackage{multirow}
\usepackage{multicol}
\usepackage{comment}
\usepackage{makecell}
\usepackage{graphicx}               
\usepackage{amsmath,amssymb}
\DeclareMathOperator*{\argmax}{arg\,max}

\newcommand{\trail}{TRAIL\xspace}

\usepackage{multirow}
\usepackage{cellspace}
\usepackage{hhline}
\usepackage{stackengine}
\setstackEOL{\\}

\begin{document}
%
\title{Learning to Guide a Saturation-Based \\ Theorem Prover}
%
%
%
%

\author{
Ibrahim Abdelaziz\thanks{$^{*}$ Equal contribution}$^*$,
Maxwell Crouse$^{*}$,
        Bassem Makni,
        Vernon Austil,
        Cristina Cornelio, 
        Shajith Ikbal,
        Pavan Kapanipathi, 
        Ndivhuwo Makondo,
        Kavitha Srinivas,
        Michael Witbrock,
        Achille Fokoue
\IEEEcompsocitemizethanks{
\IEEEcompsocthanksitem I. Abdelaziz, V. Austil, C. Cornelio, S. Ikbal, P. Kapanipathi, N. Makondo, K. Srinivas and A. Fokoue   are with IBM Research. 
\IEEEcompsocthanksitem M. Crouse is with Northwestern University
\IEEEcompsocthanksitem B. Makni is with Arizona State University
\IEEEcompsocthanksitem M. Witbrock is with The University of Auckland
\IEEEcompsocthanksitem Correspondence to Ibrahim Abdelaziz \{ibrahim.abdelaziz1@ibm.com\}, Maxwell Crouse \{mvcrouse@u.northwestern.edu\}, and Achille Fokoue \{achille@us.ibm.com\}.

}
}

\IEEEtitleabstractindextext{%
\begin{abstract}

Traditional automated theorem provers have relied on manually tuned heuristics to guide how they perform proof search. Recently, however, there has been a surge of interest in the design of learning mechanisms that can be integrated into theorem provers to improve their performance automatically.
In this work, we introduce \trail, a deep learning-based approach to theorem proving that characterizes core elements of saturation-based theorem proving within a neural framework. \trail leverages (a) an effective graph neural network for representing logical formulas, (b) a novel neural representation of the state of a saturation-based theorem prover in terms of processed clauses and available actions, and (c) a novel representation of the inference selection process as an attention-based action policy. 
We show through a systematic analysis that these components allow \trail to significantly outperform previous reinforcement learning-based theorem provers on two standard benchmark datasets (up to 36\% more theorems proved). 
In addition, to the best of our knowledge, \trail is the first reinforcement learning-based approach to exceed the performance of a state-of-the-art traditional theorem prover on a standard theorem proving benchmark (solving up to 17\% more problems).

\end{abstract}

}

\maketitle

\IEEEdisplaynontitleabstractindextext

%
\IEEEpeerreviewmaketitle


\section{Introduction}
Automated theorem provers (ATPs) are commonly used in many areas of computer science; for instance, aiding in the design of compilers \cite{curzon1991verified,leroy2009formal}, operating systems \cite{klein2009operating}, and distributed systems \cite{hawblitzel2015ironfleet,garland1998ioa}). 
As their applicability has grown in scope, there has been a need for new heuristics and strategies that inform how an ATP searches for proofs, i.e., systems that provide effective \emph{proof guidance}. Unfortunately, the specifics of when and why to use a particular proof guidance heuristic or strategy are still often hard to define \cite{schulzwe}, an issue made more concerning by how dependent theorem proving performance is on these mechanisms \cite{schulz2016performance}.



Machine learning provides one such means of circumventing this obstacle and lessening the amount of human involvement required to successfully apply an ATP to new domains. Ideally, learning-based methods would be able to automatically tune a theorem prover to the needs of a particular domain without oversight; however, in practice such methods have relied on manually designed features constructed by human experts \cite{KUV-IJCAI15-features,JU-CICM17-enigma}. Currently, there is much interest in applying deep learning to learn proof guidance strategies \cite{LoosISK-LPAR17-atp-nn,ChvaJaSU-CoRR19-enigma-ng,paliwal2019graph}, which has clear advantages with respect to the amount of feature engineering involved in their application. 

Recent neural-based approaches to proof guidance have begun to achieve impressive results, e.g., Enigma-NG \cite{JU-CoRR19-enigma-hammering} showed that purely neural proof guidance could be integrated into E prover \cite{schulz2002brainiac} to improve its performance over manually designed proof-search strategies. However, in order to achieve competitive performance with state-of-the-art ATPs, neural methods have critically relied on being seeded with proofs from an existing state-of-the-art reasoner (which itself will use a strong manually designed proof-search strategy). Thus, such approaches are still subject to the biases inherent to the theorem-proving strategies used in their initialization.

Reinforcement learning \emph{a la} AlphaGo Zero \cite{alphagozero} has been explored as a natural solution to this problem, where the system automatically learns how to navigate the search space from scratch. Examples include applying reinforcement learning to theorem proving with first-order logic  \cite{KalUMO-NeurIPS18-atp-rl,piotrowski2019guiding,zombori2019towards,zombori2020prolog}, higher-order logic \cite{BaLoRSWi-CoRR19-learning}, and also with logics less expressive than first-order logic \cite{KuYaSa-CoRR18-intuit-pl,LeRaSe-CoRR18-heuristics-atp-rl,CheTi-CoRR18-rew-rl}. 

In prior works, \emph{tabula rasa} reinforcement learning (i.e., learning from scratch) has been integrated into tableau-based theorem provers for first order logic \cite{KalUMO-NeurIPS18-atp-rl,zombori2019towards,zombori2020prolog}. Connection tableau theorem proving is an appealing setting for machine learning research because tableau calculi are straightforward and simple, allowing for concise implementations 
that can easily be extended with learning-based techniques. However, the best performing, most widely-used theorem provers to date are saturation theorem provers that implement either the resolution or superposition calculi \cite{vampire2013,schulz2002brainiac}. These provers are capable of much finer-grained management of the proof search space; however, this added power comes at the cost of increased complexity in terms of both the underlying calculi and the theorem provers themselves. 
For neural-based proof guidance to yield any improvements when integrated with highly optimized, hand-tuned saturation-based provers, it must offset the added cost of neural network evaluation with more intelligent proof search. To date, this has not been possible when these neural approaches have been trained from scratch, i.e., when they are \emph{not} bootstrapped with proofs from a state-of-the-art ATP.


In a recent work we introduced \trail \cite{crouse2019deep}, a system that could be integrated with a saturation-based theorem prover to learn effective proof guidance strategies from scratch. 
\trail demonstrated state-of-the-art performance as compared to other reinforcement learning-based theorem proving approaches on two standard benchmarks drawn from the Mizar dataset \cite{mizar40for40}: M2k \cite{KalUMO-NeurIPS18-atp-rl} and MPTP2078 \cite{urban2006mptp}. Two key aspects of \trail's design were 1) a novel mechanism used to vectorize the state of a theorem prover in terms of both inferred and provided clauses, and 2) a novel method to characterize the inference selection process in terms of an attention-based action policy. However, while most of \trail's core components were neural and thus entirely trainable, the initial vectorization scheme used to transform graph-based logical formulas into real valued vectors required hand-crafted feature extractors. Thus, while \trail made significant progress in terms of learning-based proof guidance, it was still quite limited in terms of what features it could represent and reason over.

In this work, we extend \trail in two main directions. 
First, we replace the hand-crafted feature extractors used for clause vectorization with a graph neural network designed for arbitrary logical formulas and demonstrate that this change leads to a significant improvement in performance. Second, we analyze \trail as a modular system, with a detailed analysis of its multiple components to contrast the effectiveness of each to effective theorem proving. The result of this analysis is a deeper understanding of how different components contribute to performance in learning based theorem provers such as \trail. 
In summary, our main contributions are:
\begin{itemize}
    \item An end-to-end neural approach with an efficient graph neural network vectorizer that takes into account the distinct structural characteristics of logical formulas and the time-limited nature of theorem proving.
    \item State-of-the-art performance on two standard theorem proving benchmarks as compared to other learning-based and traditional theorem provers. To the best of our knowledge, this is the first work to learn proof guidance from scratch and outperform a state-of-the-art traditional reasoner (E \cite{eprover}). 
    \item A comprehensive analysis of the performance of various components of \trail that includes different vectorization techniques, underlying reasoners, and reward structures that affect learning-based theorem proving.
    
\end{itemize}

\section{Background}\label{gen_inst}

We assume the reader has knowledge of basic first-order logic and automated theorem proving terminology and thus will only briefly describe the terms commonly seen throughout this paper. For readers interested in learning more about logical formalisms and techniques see \cite{thelogicbook,enderton2001mathematical}.

In this work, we focus on first-order logic (FOL) with equality. In the standard FOL problem-solving setting, an ATP is given a \emph{conjecture} (i.e., a formula to be proved true or false), \emph{axioms} (i.e., formulas known to be true), and \emph{inference rules} 
(i.e., rules that, based on given true formulas, allow for the derivation of new true formulas).
From these inputs, the ATP performs a \emph{proof search}, which can be characterized as the successive application of inference rules to axioms and derived formulas until a sequence of derived formulas is found that represents a \emph{proof} of the given conjecture. All formulas considered by \trail are in \emph{conjunctive normal form}. That is, they are conjunctions of {\it clauses}, which are themselves disjunctions of literals. Literals are (possibly negated) formulas that otherwise have no inner logical connectives. In addition, all variables are implicitly universally quantified.

Let $F$ be a set of formulas and $\mathcal{I}$ be a set of inference rules. We write that $F$ is \emph{saturated} with respect to $\mathcal{I}$ if every inference that can be made using axioms from $\mathcal{I}$ and premises from $F$ is also a member of $F$, i.e. $F$ is closed under inferences from $\mathcal{I}$. Saturation-based theorem provers aim to saturate a set of formulas with respect to their inference rules. To do this, they maintain two sets of clauses, referred to as the \emph{processed} and \emph{unprocessed} sets of clauses. These two sets correspond to the clauses that have and have not been yet selected for inference. The actions that saturation-based theorem provers take are referred to as \emph{inferences}. Inferences involve an inference rule (e.g. resolution, factoring) and a non-empty set of clauses, considered to be the \emph{premises} of the rule. At each step in proof search, the ATP selects an inference with premises in the unprocessed set (some premises may be part of the processed set) and executes it. This generates a new set of clauses, each of which is added to the unprocessed set. The clauses in the premises that are members of the unprocessed set are then added to the processed set. This iteration continues until a clause is generated (typically the empty clause for refutation theorem proving) that signals a proof has been found, the set of clauses is saturated, or a timeout is reached.
For more details on saturation \cite{robinson1965machine} and saturation-calculi, we refer the reader to \cite{bachmair1998equational}.

\begin{figure*}[t]
\begin{center}
\includegraphics[width=0.85\textwidth]{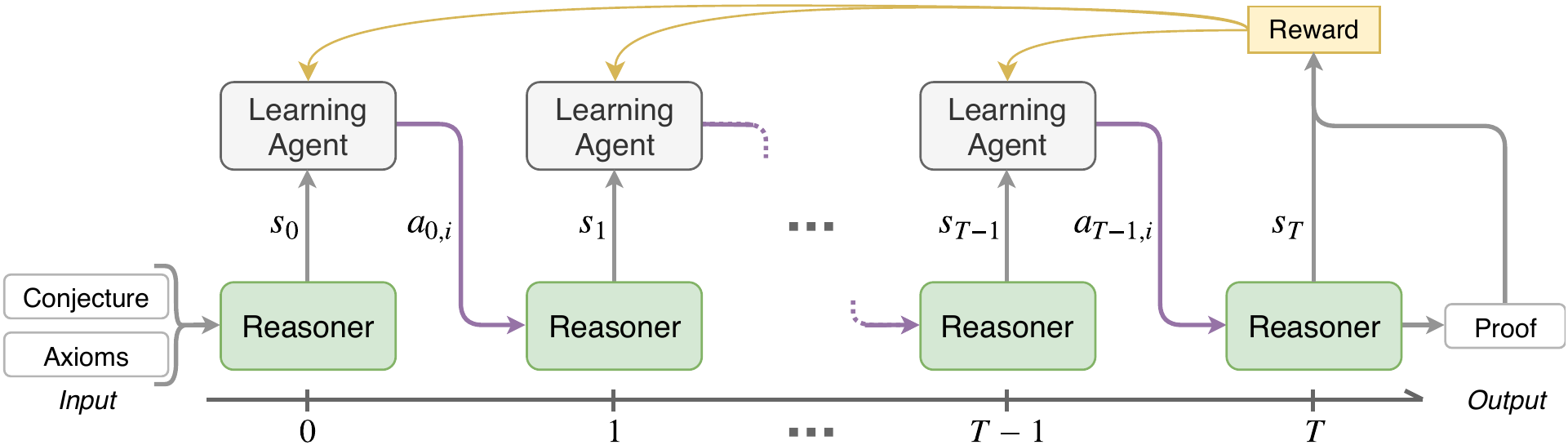}
\caption{Formulation of automated theorem proving as a RL problem}
\label{fig:rl_formulation}
\end{center}
\end{figure*}


\section{{\trail}}\label{sec:approach}
We describe next our overall approach to defining the proof guidance problem in terms of reinforcement learning. For this, we detail (a) how we formulate proof guidance as a reinforcement learning problem, (b) a \textit{neural proof state} which concisely captures the neural representations of clauses and actions within a proof state, and (c) an \textit{attention-based policy network} that learns the interactions between clauses and actions to select the next action.  Last, we describe how \trail learns from scratch, beginning with a random initial policy.

\subsection{Reinforcement Learning for Proof Guidance}
\label{sec:rl_formulation}

We formalize the proof guidance as a reinforcement learning (RL) problem where the reasoner provides the environment in which the learning agent operates.
Figure~\ref{fig:rl_formulation} shows how an ATP problem is solved in our framework.
Given a conjecture and a set of axioms, \trail iteratively performs reasoning steps until a proof is found (within a provided time limit).
The reasoner tracks the proof state, $s_t$, which encapsulates the clauses that have been derived or used in the derivation so far and the actions that can be taken by the reasoner at the current step.
At each step, this state is passed to the learning agent - an attention-based model~\cite{luong2015attention} that predicts a distribution over the actions it uses to sample a corresponding action, $a_{t,i}$. This action is given to the reasoner, which executes it and updates the proof state.

Formally, a state, $s_{t} = (\mathcal{C}_{t}, \mathcal{A}_{t})$, consists of two components. The first is the set of processed clauses, $\mathcal{C}_{t} = \{c_{t,j}\}_{j=1}^{N}$,  (i.e., all clauses selected by the agent up to step $t$); where $\mathcal{C}_{0} = \emptyset$. The second is the set of all available actions that the reasoner could execute at step $t$, $\mathcal{A}_{t} = \{a_{t,i}\}_{i=1}^{M}$; where $\mathcal{A}_{0}$ is the cross product of the set of all inference rules (denoted by $\mathcal{I}$) and the set of all axioms and the negated conjecture. An action, $a_{t,i} = (z_{t,i}, \hat{c}_{t,i})$, is a pair comprising an inference rule, $z_{t,i}$, and a clause from the unprocessed set, $\hat{c}_{t,i}$.

At step $t$, given a state $s_{t}$ (provided by the reasoner), the learning agent computes a probability distribution over the set of available actions $\mathcal{A}_t$, denoted by $P_\theta(a_{t,i} | s_{t})$ (where $\theta$ is the set of parameters for the learning agent), and samples an action $a_{t,i} \in \mathcal{A}_t$. The sampled action $a_{t,i} = (z_{t,i}, \hat{c}_{t,i})$ is executed by the reasoner by applying $z_{t,i}$ to $\hat{c}_{t,i}$ (which may involve processed clauses). This yields a set of new derived clauses, $\bar{\mathcal{C}}_{t}$, and a new state, $s_{t+1} = (\mathcal{C}_{t+1}, \mathcal{A}_{t+1})$, where $\mathcal{C}_{t+1} = \mathcal{C}_{t}\cup \{\hat{c}_{t,i}\}$ and $\mathcal{A}_{t+1} = (\mathcal{A}_{t} - \{a_{t,i}\}) \cup (\mathcal{I} \times \bar{\mathcal{C}}_{t})$.

Upon completion of a proof attempt, \trail computes a loss and issues a reward that encourages the agent to optimize for decisions leading to a successful proof in the shortest time possible. 
Specifically, for an unsuccessful  proof attempt (i.e., the underlying reasoner fails to derive a contradiction within the time limit), each step $t$ in the attempt is assigned a reward $r_t=0$. For a successful proof attempt, in the final step, the underlying reasoner produces a refutation proof $\mathcal{P}$ containing only the actions that generated derived facts directly or indirectly involved in the final contradiction. At step $t$ of a successful proof attempt where the action $a_{t, i}$ is selected, the reward $r_t$ is $0$ if $a_{t, i}$ is not part of the refutation proof $\mathcal{P}$; otherwise $r_t$ is inversely proportional to the time spent proving the conjecture. 

The final loss consists of the standard policy gradient loss~\cite{sutton1998reinforcement} and an entropy regularization term to avoid collapse onto a sub-optimal deterministic policy and to promote exploration.  
\begin{alignat*}{2}\label{eqn:loss}
       &\mathcal{L}(\theta) = &&-\mathbb{E}\big[r_t \log(P_{\theta}(a_{t} | s_{t})) \big] 
       \\ & &&-\lambda \mathbb{E}\big[\sum_{i=1}^{|\mathcal{A}_t|} - P_{\theta}(a_{t,i} | s_{t}) \log(P_{\theta}(a_{t,i} | s_{t})) \big]
\end{alignat*}
where $a_t$ is the action taken at step $t$ and $\lambda$ is the entropy regularization hyper-parameter.
This loss has the effect of giving actions that contributed to the most direct proofs for a given problem higher rewards, while dampening actions that contributed to more time consuming proofs for the same problem.

\subsection{Reward Normalization and Capping}
\label{reward_info}
As our baseline reward function, we assign a reward of 1 when a proof of the given conjecture is found and otherwise assign a reward of 0. However, because the intrinsic difficulty of problems can vary widely in our problem dataset we also explore the use of normalized rewards to improve training stability. In particular, we have implemented the following three normalization strategies (i) no normalization (baseline) (ii) normalization by the inverse of the time spent by a traditional reasoner, and (ii) normalization by the best reward obtained in repeated attempts to solve the same problem.

As noted in \cite{wang2018r}, uncapped rewards can introduce large variances when computing the gradients. To address this issue, we implemented a bounded reward structure that ensures that rewards fall within a specific range (hyper parameter) which captures how efficiently \trail solved the problem. 



\begin{figure*}
\begin{center}
\includegraphics[width=0.9\textwidth]{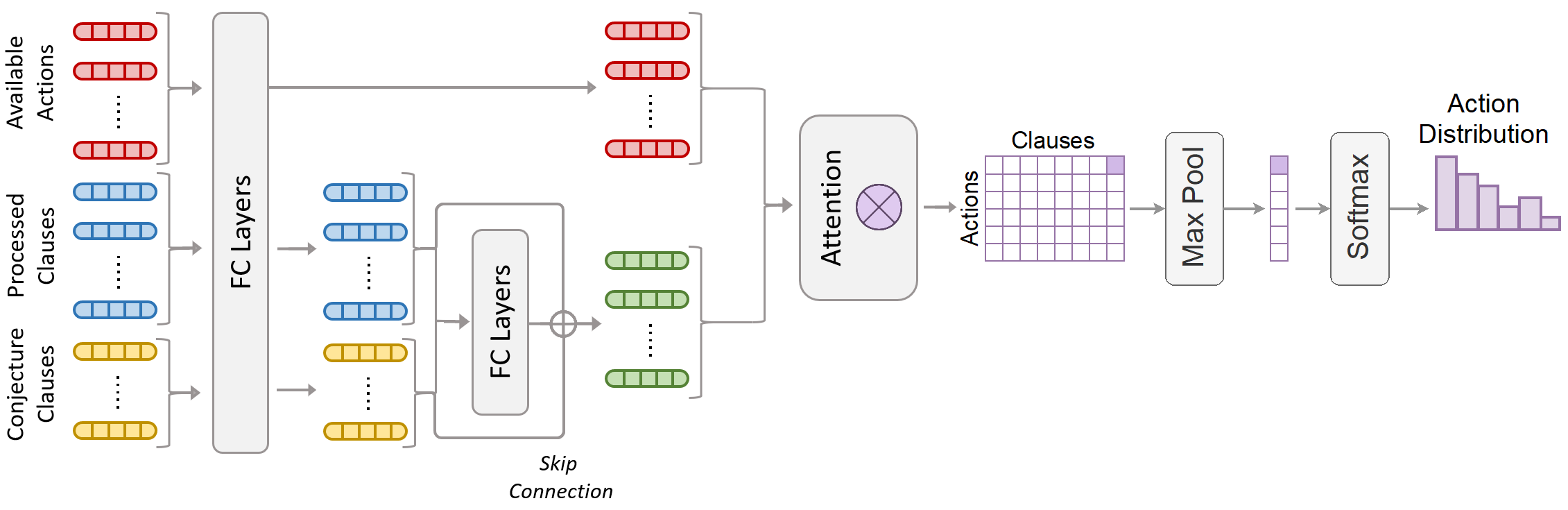}
\end{center}
\caption{Flow from clause vectorization through the policy network}
\label{fig:network_arch}
\end{figure*}

\subsection{Neural Representation of Proof State}

Recall that the proof state consists of the sets of processed clauses $\mathcal{C}_t$ and actions $\mathcal{A}_t$. Each clause in the processed set and in the set of available actions is first transformed into a vector representation (this transformation will be described in Section~\ref{sec:vectorization}).
To produce dense representations for the elements of $\mathcal{C}_t$ and $\mathcal{A}_t$, \trail then passes these clause representations 
through $k$ fully-connected layers. This yields sets $\{\mathbf{h}^{(p)}_{t,1},..., \mathbf{h}^{(p)}_{t,N}\}$ and $\{\mathbf{h}^{(a)}_{t,1},..., \mathbf{h}^{(a)}_{t,M}\}$ of dense representations for the processed and action clauses. \trail also collects the dense representations for the negated conjecture clauses as $\{\mathbf{h}^{(c)}_{1}, \ldots \mathbf{h}^{(c)}_k\}$.

Following this transformation, the dense representations for each action are collected into a matrix $\mathbf{A}$.
To construct the processed clause matrix, \trail first produces a dense representation of the conjecture as the element-wise mean of dense negated conjecture clause representations
\begin{align*}
\mathbf{h}^{(c)} = \dfrac{1}{k}\sum_{i = 1}^k \mathbf{h}^{(c)}_i
\end{align*}
where $k$ is the number of conjecture clauses. New processed clause representations are produced by combining the original dense representations with the pooled conjecture. For a processed clause embedding $\mathbf{h}^{(p)}_i$, its new value would be
\begin{align*}
\hat{\mathbf{h}}^{(p)}_i = \mathbf{h}^{(p)}_i + \mathbf{h}^{(c)} + F(\mathbf{h}^{(p)}_i || \  \mathbf{h}^{(c)})
\end{align*}
where \textit{F} is a feed-forward network, $||$ denotes the concatenation operation, and the original inputs are included with skip connections \cite{he2016deep}. The new processed clause embeddings are then joined into a matrix $\mathbf{C}$.

The two resulting matrices $\mathbf{C}$ and $\mathbf{A}$ can be considered the neural forms of $\mathcal{C}_t$ and $\mathcal{A}_t$. Thus, they concisely capture the notion of a \emph{neural proof state}, where each column of either matrix corresponds to an element from the formal proof state. Following the construction of $\mathbf{C}$ and $\mathbf{A}$, this neural proof state is fed into the policy network to select the next inference.

\subsection{Attention-Based Policy Network}\label{sec:policy-network}

Figure~\ref{fig:network_arch} shows how clause representations are transformed into the neural proof state and passed to the policy network. Throughout the reasoning process, the policy network must produce a distribution over the actions relative to the clauses that have been selected up to the current step, where both the actions and clauses are sets of variable length.
This setting is analogous to ones seen in attention-based approaches to problems like machine translation~\cite{luong2015attention,vaswani2017attention} and video captioning~\cite{yu2016video,whitehead2018kavd}, in which the model must score each encoder state with respect to a decoder state or other encoder states.
To score each action relative to each clause, we compute a multiplicative attention~\cite{luong2015attention} as
 \begin{equation*}
 \mathbf{H} = \mathbf{A}^\top \mathbf{W}_a \mathbf{C} ~, 
 \end{equation*}
where $\mathbf{W}_a \in \mathbb{R}^{(2d + |\mathcal{I}|) \times 2d}$ is a learned parameter and the resulting matrix, $\mathbf{H} \in \mathbb{R}^{M \times N}$, is a heat map of interaction scores between processed clauses and available actions. \trail then performs max pooling over the columns (i.e., clauses) of $\mathbf{H}$ to produce unnormalized action values $\widehat{P}_\theta(a_{t,i} | s_{t})$

Prior work integrating deep learning with saturation-based ATPs would use a neural network to score the unprocessed clauses with respect to \emph{only} the conjecture and \emph{not} the processed clauses \cite{LoosISK-LPAR17-atp-nn,JU-CoRR19-enigma-hammering}. \trail's attention mechanism can be viewed as a natural generalization of this, where inference selection takes into account both the processed clauses and conjecture.


\subsection{Learning From Scratch}\label{sec:train}

\trail begins learning through random exploration of the search space as done in AlphaZero \cite{alphazero} to establish performance when the system is started from a \textit{tabula rasa} state (i.e., a randomly initialized policy network $P_\theta$).  At training, at an early step $t$ (i.e., $t < \tau_0$, where $\tau_0$, the temperature threshold, is a hyper-parameter that indicates the depth in the reasoning process at which training exploration stops), we sample the action $a_{t,i}$ in the set of available actions  $\mathcal{A}_{t}$, according to the following probability distribution $P_\theta$ derived from the policy network's output $\widehat{P}_\theta$:
\begin{align*}
    P_\theta(a_{t,i} | s_{t}) = \frac{\widehat{P}_\theta(a_{t,i} | s_{t})^{1/\tau}}{\mathlarger\sum_{a_{t,j} \in \mathcal{A}_{t} }{\widehat{P}_\theta(a_{t,j} | s_{t})^{1/\tau}} }
\end{align*}
where $\tau$, the temperature, is a hyperparameter that controls the exploration-exploitation trade-off and decays over the iterations (a higher temperature promotes more exploration). When the number of steps already performed is above the temperature threshold (i.e., $t \geq \tau_0$), an action, $a_{t,i}$, with the highest probability from the policy network, is selected, i.e.
\begin{align*}
a_{t,i} = \argmax_{a_{t,j} \in \mathcal{A}_{t} } P_\theta(a_{t,j} | s_{t}) 
\end{align*}    
At the end of training iteration $k$, the newly collected examples and those collected in the previous $w$ iterations ($w$ is the example buffer hyperparameter) are used to train, in a supervised manner, the policy network using the reward structure and loss function defined in Section~\ref{sec:rl_formulation}.

\section{Logical Formula Vectorization}
\label{sec:vectorization}
For the policy network to reason over a theorem prover's state, \trail must transform its internal graph-based logical formulas into real-valued vectors.
To do this, \trail utilizes a set of $M$ vectorization modules, $\mathcal{M} = \{ m_1, \ldots, m_M \}$, that each characterize some important aspect of the clauses and actions under consideration. Letting $m_k(i)$ be the vector for an input clause $i$ produced by module $m_k \in \mathcal{M}$, the final vector representation $v_i$ is then the concatenation of the outputs from each module. This combined vector representation is passed through a set of fully-connected layers and then sent to the policy network, as Figure~\ref{fig:network_arch} shows. Within the set of all vectorization modules, we broadly distinguish between modules that represent \emph{simple} and \emph{structural} characteristics of their input clauses. 


\subsection{Capturing Simple Features}

The first set of vectorization modules considered are those capturing simple features that may not be readily inferrable from their input clauses. Currently, the set of features represented are the age (the timestep at which a clause was derived), weight (the number of symbols in the clause), literal count, and set-of-support (whether or not the clause has a negated conjecture clause as an ancestor in its proof tree). Each such vectorization module $m_k \in \mathcal{M}$ follows the same general design. Given an input clause, $m_k$ produces a discrete, bag-of-words style vector in $\mathbb{Z}^{n_k}$, where $n_{k}$ is a pre-specified dimensionality specific to module $m_k$. As an example, consider the module intended to represent a clause's age (i.e., at what timestep the clause was generated). It would map a clause to a one-hot vector where the only non-zero index is the index associated with the clause's age (ranging from $1, \ldots, n_k$, where $n_k$ is the maximum tracked age).

\subsection{Capturing Structural Features}

In addition to the vectorization modules presented above, \trail also incorporates modules designed to capture the complex structural information inherent to logical formulas. In the previous iteration of \trail \cite{crouse2019deep}, this included Enigma \cite{JU-CICM17-enigma} modules which characterized a clause in terms of fixed-length term walks (with separate modules for different lengths) and a chain-pattern module that extracted linear chains between the root and leaf nodes of a literal.

The time-sensitive nature of theorem proving makes simple, fast-to-extract features like those previously mentioned quite appealing. However, such features generally carry strong assumptions as to what is and is not useful to proof search that may ultimately limit the invention of new, more effective strategies. As part of a shift towards more learned representations, the pattern-based feature extractors have been replaced with graph neural network-based modules. Graph neural networks have been proposed as a means of overcoming the hand-engineering required by previous feature extraction methods, however their application within the internal operations of a theorem prover is still relatively recent \cite{rawsonreinforced,paliwal2020graph}.

Initially, \trail used only off-the-shelf implementations of two popular GNN architectures, the GCN~\cite{kipf2017gcn} and GraphSAGE~\cite{hamilton2017inductive}. However, neither of these methods fully leveraged the unique characteristics of logical formulas. This motivated the development of a new method, which we refer to as the StagedGCN, that takes into account this information in its embedding of a logical formula while also being conscious of the time-limited nature of theorem proving. We describe each of these architectures in the rest of this section.


\subsubsection{Logical Formulas as Graphs}
\begin{figure}
\begin{center}
\includegraphics[width=0.5\columnwidth] {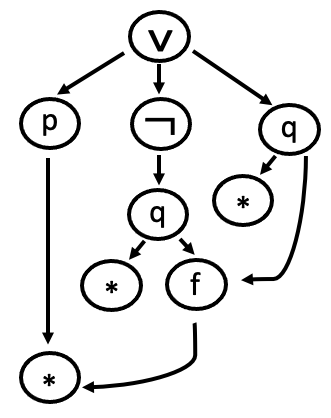}
\caption{DAG representation of the clause $\forall A, B, C .\ p(A) \vee \lnot q(B, f(A)) \vee q(C, f(A))$}
\label{fig:dag_example}
\end{center}
\end{figure}

As in recent works ~\cite{wang2017premise, paliwal2020graph}, we represent logical formulae as directed acyclic graphs (DAGs), illustrated in Figure~\ref{fig:dag_example}. This is achieved by taking the parse tree of a formula and merging together any nodes that correspond to identical subtrees. In addition, variables have their labels replaced with generic tokens to ensure the resultant graph is invariant to arbitrary variable renamings.

\subsubsection{Graph Convolution Neural Networks}
Graph convolutional neural networks (GCNs) \cite{kipf2017gcn}  compute the embedding of a node as an aggregation of the embeddings for its neighbor nodes.
\begin{alignat*}{2}
h_{u}^{(i)} &= \sigma \big( W^{(i)} \big( \dfrac{h_u^{(i - 1)}}{|\mathcal{N}(u)|} + \sum_{v \in \mathcal{N}(u)} \dfrac{h_{v}^{(i - 1)}}{\sqrt{|\mathcal{N}(u)||\mathcal{N}(v)|}} \big) \big)
\end{alignat*}
where $\sigma$ is a non-linearity (in this work, we use a ReLU), $\mathcal{N}(u)$ is the set of neighbors for node $u$, and $W^{(i)}$ is a learned matrix specific to the $i$-th round of updates. Following $K$ rounds of updates, the embeddings for each node are first collected into a matrix $\mathbf{H}$ and then a pooling operation is applied to generate a single vector $\mathbf{h}$ for the input graph.

\subsubsection{GraphSAGE}

The GraphSAGE architecture of \cite{hamilton2017inductive} is a generalization of the GCN that allows for trainable aggregation operations. In this work, the aggregation is based on mean-pooling of neighbors to a given node
\begin{alignat*}{2}
&\hat{h}^{(i)}_u &&= \sigma\big(\textit{W}^{(i)}_A \big( \dfrac{1}{|1 + \mathcal{N}(u)|}\sum_{v \in \mathcal{N}(u) \cup \{u\}}h^{(i-1)}_v \big) \big) \\
&h^{(i)}_u &&= \sigma\big(\textit{W}^{(i)} (h^{(i - 1)}_u \ || \ \hat{h}^{(i)}_u)\big)
\end{alignat*}
where $||$ denotes vector concatenation, $\textit{W}^{(i)}_A$  and $\textit{W}^{(i)}$ denote weight matrices specific to layer $i$, and $\sigma$ is a ReLU non-linearity. Like the GCN, the node update step is performed for each node for $K$ rounds of updates, with the final set of node embeddings being collected into the matrix $\mathbf{H}$ and then pooled to produce $\mathbf{h}$.

\begin{figure}
\begin{center}
\includegraphics[width=.5\columnwidth] {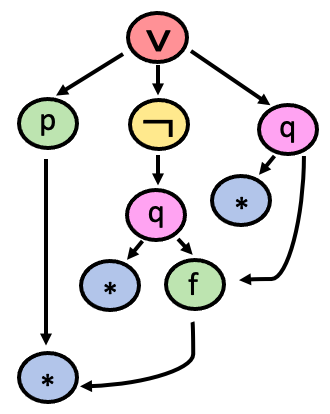}
\caption{Update staging in StagedGCN. Identically colored nodes are updated simultaneously}
\label{fig:topo_batch}
\end{center}
\end{figure}



\subsubsection{StagedGCN: Graph Convolution Neural Networks With Topological Batching}
The GCN presented earlier does not attempt to capitalize on the unique DAG structure of  logical formulas. We therefore introduce a new graph embedding method derived from Relational GCNs (R-GCNs) \cite{schlichtkrull2018modeling}. This method is referred to as a \emph{StagedGCN}, and is inspired from recent works~\cite{crouse2019improving, thost2021directed} that leverage directed and acyclic structure by processing, evaluating, and updating node embeddings according to a topological sort of the graph. As shown in Figure~\ref{fig:topo_batch}, node updates start from the leaf nodes (blue nodes), then proceed to green nodes, pink nodes, yellow nodes, and finally red nodes. Nodes that do not share any dependencies (shown in the figure as nodes with the same color) can be updated simultaneously. That is, to boost efficiency we may batch together the updates of such nodes. Information propagates from leaf nodes to the root node. The update function is as follows:
\begin{align*}
\hat{h}_{u}^{(i)} &= \sigma\big(  
\textrm{LN}\big(c_{u}W^{(i)} h_{u}^{(i-1)} +
\sum_{r\in \mathcal{R}}
\sum_{v\in\mathcal{N}_{u,r}} 
{c_{u,r}}W_{r}^{(i)} h_{v}^{(i)} \big)\big) \\
h_{u}^{(i)} &= \hat{h}_{u}^{(i)} + h_{u}^{(i-1)} 
\end{align*}
Here, $\mathcal{R}$ is the set of edge types; 
$\mathcal{N}_{u,r}$ is the set of neighbors connected to node $u$ through the edge type $r$; $c_{u}$ and $c_{u,r}$ are  normalization constants;
$W_{r}^{(i)}$ are the learnable weight matrices, one per $r\in \mathcal{R}$; $W^{(i)}$ is a learnable weight matrix; 
$\sigma$ is a non-linear activation function (here tanh); and $\textrm{LN}$ is the layer normalization function~\cite{ba2016layer}. Importantly, there are two differences that must be emphasized in the above update equations for the StagedGCN as compared to the relational GCN:
\begin{enumerate}
    \item In the computation of the intermediary embedding $\hat{h}_{u}^{(i)}$ of node $u$,  we use the node embeddings $h_{v}^{(i)}$ of neighbors $v$ of $u$ in the current iteration $i$ whereas standard relational GCN uses the embeddings $h_{v}^{(i-1)}$ from the previous iteration $i-1$. That computation also explicitly uses the node embedding of $u$ at the previous iteration $i-1$ (i.e., $ h_{u}^{(i-1)} $).
    \item We use a residual connection as shown in the second update equation: $h_{u}^{(i)} = \hat{h}_{u}^{(i)} + h_{u}^{(i-1)}$
\end{enumerate}

For a graph $G$, to also take into account information propagation from the root to leaf nodes, we use a bidirectional embedding as follows:
\begin{alignat*}{2}
h_{u}^{(i)}(G) &= F_{BD}\big( [ h_{u}^{(i)}(G^{\uparrow}) \ || \ h_{u}^{(i)}(G^{\downarrow})] \big)
\end{alignat*}
where $F_{BD}$ is a feed-forward network with residual connections; $||$ is the concatenation operator; $G^{\downarrow}$ is the graph G with the direction of its edges reversed; $G^{\uparrow}$ is the graph G with the direction of its edges unchanged; and $ h_{u}^{(i)}(G^{\uparrow})$ and $h_{u}^{(i)}(G^{\downarrow})$ are updated embeddings of node $u$ at iteration $i$ (following the design presented above), which accumulate node embeddings towards the root and children, respectively. 

After $K$ update iterations, the final embedding of the rooted directed acyclic graph G is given as the following function of the final embedding of its root node $root$ only: 
$$\rho(\textrm{LN}(\overline{W}h_{root}^{(K)}(G)))$$
where $\overline{W}$ is a learnable weight matrix and $\rho$ is an activation function (here ReLU).


\section{Experiment Setup}
\label{sec:experiment_setup}

\subsection{Datasets}
We use two standard, benchmark datasets to evaluate \trail's effectiveness: \emph{M2k}~\cite{KalUMO-NeurIPS18-atp-rl} and \emph{MPTP2078}~\cite{alama2014premise} (referenced as MPTP going forward). Both M2k and MPTP are exports of parts of Mizar\footnote{\url{https://github.com/JUrban/deepmath/}}~\cite{mizar} into the TPTP~\cite{tptp} format. The M2k dataset contains 2003 problems selected randomly from the subset of Mizar that is known to be provable by existing ATPs, whereas MPTP is a set of 2078 Mizar problems selected regardless of whether or not they could be solved by an ATP system. 

\begin{table}[]
    \caption{Hyperparameter values }
\centering
\begin{tabular}{ll}
\toprule
Hyperparameter & Value   \\ 
\midrule
    $\tau$ (temp.)              &   3       \\  
    $\tau_0$ (temp. threshold)  &   11000  \\
    Dropout                     &   0.57     \\
 Learning rate               &   0.001   \\
    Temp. decay                  &   0.89.   \\  
    $\lambda$ (reg.)            &   0.004 \\    
    Epochs                  &  10     \\
    Min reward              &  1.0       \\   
    Max reward              &   2.0     \\  
    Activation function &  ReLU    \\
    Reward normalization        &   normalized    \\ 
    GCN conv. layers        &  2     \\
\bottomrule
\end{tabular}
    \label{tab:hyperparams_tunable}%
\end{table}

\begin{table*}[t]
\caption{Number of problems solved in M2k and MPTP2078. Numbers for learning-based approaches are for best iteration/cumulative. Best two approaches in \bf{bold}. 
}
\centering
\begin{tabular}{@{}llcc|cc@{}}
\toprule
& & \multicolumn{2}{|c|}{MPTP} & \multicolumn{2}{c}{M2k} \\ 
  \cmidrule(l){3-6} & & \multicolumn{1}{|l}{Best Iteration} & \multicolumn{1}{l|}{Cumulative} & \multicolumn{1}{l}{Best Iteration} & \multicolumn{1}{l}{Cumulative} \\ 
 \midrule
 \multirow{3}{*} {\setstackgap{S}{4.05ex}\Centerstack[l]{Learning-Based}} & \multicolumn{1}{l|}{rlCop} & 622 & 657 & 1,253 & 1,350 \\
& \multicolumn{1}{l|}{plCop} & 860 & 890 & 1,372 & 1,416             \\
& \multicolumn{1}{l|}{\trail} & \bf{1,167} & \textbf{1,213} & \bf{1,776} & \textbf{1,808}              \\ 
\midrule
\midrule
\multirow{3}{*} {\setstackgap{S}{4.05ex}\Centerstack[l]{Traditional}} & \multicolumn{1}{l|}{E} &  \multicolumn{2}{c|}{\bf 1,036} & \multicolumn{2}{c}{\bf 1,819} \\
& \multicolumn{1}{l|}{Beagle} & \multicolumn{2}{c|}{742} &  \multicolumn{2}{c}{1,543} \\
& \multicolumn{1}{l|}{mlCop} &  \multicolumn{2}{c|}{502} &  \multicolumn{2}{c}{1,034} \\
\bottomrule
\end{tabular}
\label{tab:m2k_2078}
\end{table*}

\subsection{Hyperparameter Tuning and Experimental Setup}
We  used  gradient-boosted tree  search from
scikit-optimize\footnote{\url{https://scikit-optimize.github.io/}} to  find  effective  hyperparameters using 10\% of the Mizar dataset\footnote{We ensured that the 10\% used for hyperparameter tuning did not overlap with any of MPTP or M2k}. This returned the hyperparameter values in Table~\ref{tab:hyperparams_tunable}. Experiments were conducted over a cluster of 19 CPU (56 x 2.0 GHz cores \& 247 GB RAM) and 10 GPU machines (2 x P100 GPU, 16 x 2.0 GHz CPU cores, \& 120 GB RAM) over 4 to 5 days (for hyperparameter tuning, we added 5 CPU and 2 GPU machines).

All experiments for TRAIL and all its competitors are run with a maximum of 100 seconds time limit per problems. Furthermore, we set the number of iterations for TRAIL and learning-based approaches (rlCop~\cite{KalUMO-NeurIPS18-atp-rl} and plCop~\cite{zombori2020prolog}) to 20 iterations. We use two metrics to measure performance. The first is \textit{cumulative completion performance} which, following \cite{BaLoRSWi-CoRR19-learning}, is the cumulative number of distinct problems solved by \trail across all iterations. The second metric is \textit{best iteration completion performance}. This was reported in \cite{KalUMO-NeurIPS18-atp-rl} and is the number of problems solved at the best performing iteration.

\subsection{Reasoner Integration}
The current implementation of \trail assumes only
that its underlying reasoner is saturation-based and is otherwise reasoner-agnostic. It defines an API that can be implemented by any reasoner to allow \trail to act as its proof guidance system. 
The API includes a set of functions needed for TRAIL to guide the reasoner in a client-server fashion. For each problem, TRAIL first initialize the reasoner with the problem to be proved. Then it repeats the following steps in a loop until either proof status turns true or the time limit is reached: 1) request reasoner for the current state and proof status 2) estimate the next action to be executed, 3) request reasoner to execute the next action.
Throughout all experiments presented in the following sections, the reasoner in use has its own proof guidance completely disabled when integrated with \trail. Whenever the reasoner reaches a decision point, it delegates the decision to \trail's policy to choose the next action to pick. Using an off-the-shelf reasoner (like Beagle~\cite{Beagle2015} or E-prover~\cite{eprover}) to execute inferences ensures that the set of inference rules available to \trail are both sound and complete, and that all proofs generated can be trusted. For all experiments in this work, \trail is integrated with E-prover.

\section{Experiments and Results}
\subsection{Effectiveness of TRAIL}
For our main results, we compare the raw performance of \trail against previous theorem proving approaches. Following \cite{BaLoRSWi-CoRR19-learning,KalUMO-NeurIPS18-atp-rl,zombori2020prolog}, we report the number of problems solved at best iteration and the number solved cumulatively across iterations. In this setting, \trail starts from scratch (i.e., from random initialization) and is applied to M2k and MPTP for 20 iterations, with learning from solved problems occurring after each iteration completes. 
As with \cite{zombori2020prolog}, we limit TRAIL to a maximum of 2,000 steps per problem with a hard limit of 100 seconds. 

For traditional ATP systems, we compare \trail to: 
\begin{enumerate}
    \item \textbf{E}~\cite{schulz2002brainiac} in auto mode, a state-of-the-art saturation-based ATP system that has been under development for over two decades
    \item \textbf{Beagle}~\cite{Beagle2015}, a newer saturation-based theorem prover that has achieved promising results in recent ATP competitions
    \item \textbf{mlCop} \cite{kaliszyk2015certified}, an OCaml reimplementation of leanCop \cite{otten2003leancop}, which is a tableau-based theorem prover that was applied to M2k in \cite{KalUMO-NeurIPS18-atp-rl} and MPTP in \cite{zombori2020prolog}
\end{enumerate}
For learning-based approaches, we compare against two recent RL-based systems: 4) \textbf{rlCop} ~\cite{KalUMO-NeurIPS18-atp-rl} and 5) \textbf{plCop}~\cite{zombori2020prolog}, both of which are connection tableau-based theorem provers that build off mlCop and leanCop, respectively. 



\begin{table*}[t]
\caption{Performance of different vectorizers in terms of problems solved. The number in brackets denotes the number of problems solved by TRAIL and not by E.}
\centering
\begin{tabular}{@{}lcc|cc@{}}
\toprule
 & \multicolumn{2}{|c|}{MPTP}                                             & \multicolumn{2}{c}{M2k}                                              \\ \cmidrule(l){2-5} 
    & \multicolumn{1}{|l}{Best Iteration} & \multicolumn{1}{l|}{Cumulative} & \multicolumn{1}{l}{Best Iteration} & \multicolumn{1}{l}{Cumulative} \\ \midrule
\multicolumn{1}{l|}{StagedGCN} & \textbf{1,167}                               & \textbf{1,213 (\small 228)}              & \textbf{1,776}                               & \textbf{1,808 (\small 50)}              \\
\multicolumn{1}{l|}{GCN}              & 892                                & 998 (\small 101)                        & 1,610                               & 1,684 (\small 23)                       \\
\multicolumn{1}{l|}{SAGE}                       & 1,061                               & 1,150 (\small 186)                       & 1,678                               & 1,739 (\small 36)                       \\
\multicolumn{1}{l|}{Pattern-based}            & 1,095                               & 1,155 (\small 185)                       & 1,732                               & 1,765 (\small 46)                       \\
\bottomrule
\end{tabular}
\label{sparse_vs_neural}
\end{table*}

Table~\ref{tab:m2k_2078} shows the performance of \trail against traditional and learning-based approaches. Compared to RL-based approaches plCop~\cite{zombori2020prolog} and rlCop~\cite{KalUMO-NeurIPS18-atp-rl}, \trail achieves significantly better performance on both benchmarks. On the M2k benchmark, \trail solved cumulatively 1,808 problems compared to 1,416 and 1,350 for plCop and rlCop (an improvement of 19\% - 22\%). Similarly, on MPTP, \trail solves 1,213 problems in total  where plCop and rlCop solved only 890 and 657, respectively. 
\trail is designed for saturation-based theorem provers, which are known generally to be more effective than the tableau-based provers against which we compare (largely due to their more effective control of the proof search space). Thus, TRAIL gets the benefits of both saturation calculi (which use powerful term orderings, literal selection techniques, and redundancy elimination rules) as well as the benefits of our new more powerful proof guidance heuristics (through the learned neural modules). 

The performance gain of \trail is also clear when compared to traditional non-learning theorem provers. Specifically, \trail significantly outperforms mlCop and Beagle on both M2K (1808 versus 1543 and 1034) and MPTP (1213 versus 742 and 502). Compared to the state-of-the-art prover, E, \trail achieves very comparable performance on M2K dataset (1808 by \trail against 1819 by E). However, on the dataset containing problems not yet solvable by traditional theorem provers, MPTP, \trail achieves much better performance as compared to E with \trail solving 1213 problems and E solving 1036.

Interestingly, out of the 1213 MPTP problems solved by \trail, 218 were never solved by E within the specified time limit. Further, even on M2k, where both \trail and E are near the limit of solvable problems ($\sim$1800 out of 2003), \trail solved 49 problems that E did not solve and E solved 60 that \trail did not solve. These differences suggest that \trail is \emph{not} simply learning E's proof search strategy. 

Overall, these results show that \trail (when trained from a random initial policy) is a competitive theorem proving approach with demonstrated improvements over all existing approaches. In the rest of this section, we analyze the performance of \trail further to show the impact of its different design decision such as reward structure, choice of vectorizer and a qualitative performance.

\subsection{Impact of Vectorization Strategy}

The original implementation of \trail \cite{crouse2019deep} represented logical formulas as sets of cheap-to-extract patterns that could be trivially converted into sparse vector representations to serve as input for the policy network. This was advantageous from a computational efficiency perspective, however, it imposed strong inductive biases on the learning process that may have ultimately limited \trail's problem solving capabilities. In this section, we compare the impact that various formula vectorization techniques have on theorem proving performance. In particular, we compare \trail's current formula vectorization scheme to alternative graph neural network-based vectorization methods and the pattern-based vectorizer previously used by \trail, including a detailed analysis of the results. 

\subsubsection{Performance}
Table~\ref{sparse_vs_neural} shows the performance of \trail using 4 different  vectorizers on the the two datasets. For each vectorizer, we report both best iteration performance as well as cumulative completion performance. The numbers in parentheses indicate the number of problems solved by \trail with this vectorizer which E-prover  could not solve. 

As can be seen in the table, the vanilla off-the-shelf GCN vectorizer solved the fewest problems in both datasets. The more advanced SAGE architecture did much better in terms number of the overall problems solved (1150 on MPTP and 1739 on M2k) and how many problems it solved that E did not (168 on MPTP and 36 on M2K). The pattern-based vectorizer slightly outperformed SAGE, likely due to its computational efficiency.  
The StagedGCN vectorizer (which better accounts for the unique characteristics of logical formulas) performed the best among all vectorizers tested. 

\begin{table}
\caption{Performance when models are trained and tested on distinct datasets}
\centering
\begin{tabular}{lcc}
\toprule
     & \multicolumn{1}{l}{MPTP $\rightarrow$ M2K} & \multicolumn{1}{l}{M2K $\rightarrow$ MPTP}    \\
\midrule
StagedGCN & \bf 1,674     & \bf 1,048 \\
GCN & 1,602 & 732 \\
SAGE  & 1,649    & 871 \\
Pattern-based   & 1,623      & 969    \\
\bottomrule
\end{tabular}
\label{tab:transfer}
\end{table}

\subsubsection{Generalizability}
The standard evaluation scheme (wherein one measures performance across iterations on the same set of problems) does not adequately demonstrate how effective a trained system would be in generalizing to an unseen set of problems. For our next experiment we assessed the performance of \trail using different vectorization techniques in terms of generalization. As both MPTP and M2k were drawn from the same Mizar library of problems, this experiment simply involved training each vectorizer on one dataset and testing on the other (with overlapping test problems being filtered out). Table~\ref{tab:transfer} shows the performance of TRAIL under this evaluation scheme for each vectorizer. Interestingly, each vectorizer seemed to achieve nearly the same performance as they had under the standard iterative evaluation (with the StagedGCN again producing the best results). This provides evidence that the representations \trail learns with each vectorization method are reasonably generalizable.

\begin{figure*}[h]
\begin{subfigure}
  \centering
  \includegraphics[width=.48\linewidth]{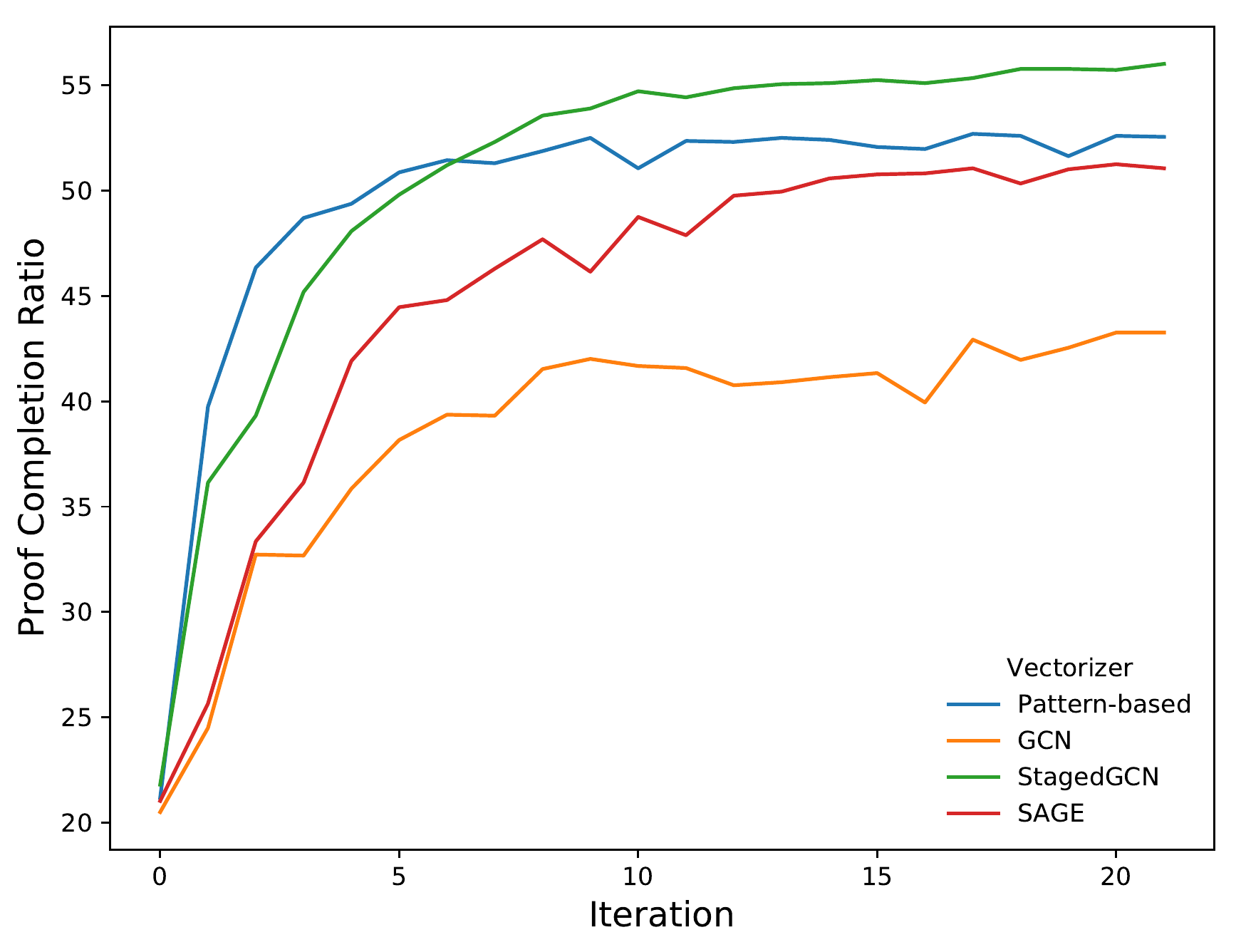}  
  \label{fig:sub-first}
\end{subfigure}\hfill
\label{fig:2078b_compl_time}
\begin{subfigure}
  \centering
  \includegraphics[width=.48\linewidth]{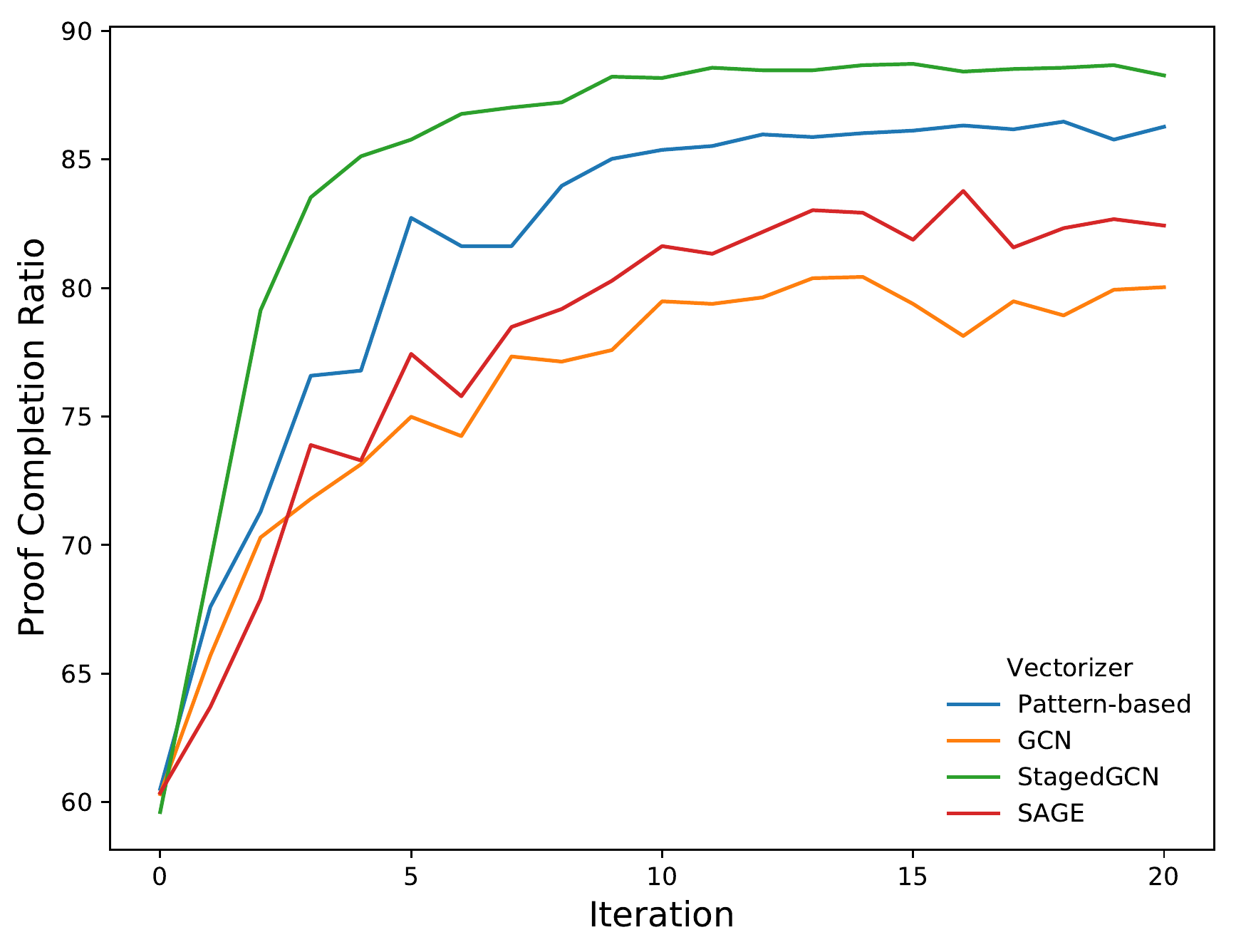}  
  \label{fig:sub-first}
\end{subfigure}\hfill
\caption{Completion ratio for different vectorizers on MPTP (left graph) and M2k (right graph)}
\label{fig:m2k_compl_time}
\end{figure*}

\subsubsection{Completion Ratio}
The graphs in Figure \ref{fig:m2k_compl_time} show the completion ratio for all vectorizers across iterations. The completion ratio captures the percentage of problems solved in each dataset per iteration (e.g., if \trail solved 980 out of the 2078 MPTP problems, this value would be $47\%$).

The left graph in Figure \ref{fig:m2k_compl_time} shows the completion ratio across iterations on the MPTP benchmark. There we can see that pattern-based vectorization was superior in the first 6 iterations. This can likely be attributed to both its computational efficiency (pattern-based features are cheap to extract) as well as its built-in domain knowledge. At iteration 7, the StagedGCN overtook the pattern-based vectorizer and then remained as the top performer for the remaining iterations. This supports the notion that the hand-engineered pattern-based features, while useful for theorem proving, could not capture the complete set of features needed to best guide proof search. In the right graph, we can see that the StagedGCN was uniformly better than all other vectorizers at each iteration. It is unclear why the StagedGCN outperformed the pattern-based vectorizer in the earlier iterations; however, we suspect that this is due to the comparatively easier difficulty of the M2k dataset. That is, far more problems were solvable at each iteration than in MPTP, and thus there was more data with which to train the more complex GNN models.  

\subsubsection{Vectorization Strategy Conclusions}
As shown in our experiments, the StagedGCN vectorization method introduced in this work provided the most utility for \trail on both benchmarks. This is evident when comparing the convergence in terms of the number of problems solved for both datasets. In terms of generalization, each vectorizer showed reasonable ability, with all methods achieving roughly the same performance as in the standard iterative evaluation setting. Overall, we thus conclude that the StagedGCN is the best vectorization method. It provides the best performance, and does not appear to have any clear downsides as compared to the alternative vectorizers.

\subsection{StagedGCN Ablation Study}

\noindent \textit{Node Embedding Size:} In this experiment, we vary the embedding size for the best performing vectorizer; the StagedGCN. In particular, we vary clause and action embedding sizes between 64, 128, 256 and 512. Table~\ref{tab:embed_size} shows the performance on both datasets. Lower embedding dimensions are preferred in our setting, since they lead to fewer weights and thus faster training and inference times. Table~\ref{tab:embed_size} clearly shows this, with embedding sizes 64 and 128 outperforming embedding sizes of 256 and 512 on MPTP (with 64 size being the best overall across both datasets). 



\begin{table}
\centering
\begin{subtable}{}
\caption{StagedGCN performance (cumultative problems solved) across various clause embedding sizes}
\centering
\begin{tabular}{lll}
\toprule
 Embedding Size   & MPTP & M2k  \\
\midrule
512 & 1,199  & 1,800 \\
256 & 1,189  & 1,807 \\
128 & \bf 1,215  & 1,800 \\
64  & 1,213  & \bf 1,808\\
\bottomrule
\end{tabular}%
\label{tab:embed_size}
\end{subtable}
\\
\begin{subtable}{}
\caption{Ablation study of StagedGCN comparing cumulative number of problems solved}
\centering
\begin{tabular}{lcc}
\toprule
   & MPTP & M2k  \\
\midrule
TRAIL (No ablation) & \textbf{1,213}  & \textbf{1,808} \\
No Layer Normalization & 799   &  1,462\\
No Skip Connections & 1,129  & 1,707 \\
No Root Readout &  1,212  & 1,805 \\
\bottomrule
\end{tabular}%
\label{tab:ablation}
\end{subtable}

\begin{subtable}{}
\caption{Performance (cumulative problems solved) when varying reward sources, reward normalization, bounding rewards, and exploration}
\centering
\begin{tabular}{lcccc}
\toprule
    Source & Normalized? & Bounded? & Exploration? 
    & MPTP  \\
\midrule
Time-based & \checkmark & \checkmark & \checkmark & \bf 1,213 \\
Time-based & -- & \checkmark & \checkmark & 1,197 \\
Step-based & \checkmark & \checkmark & \checkmark & 1,192 \\
Step-based & -- & \checkmark & \checkmark & 1,167 \\
Time-based & \checkmark & \checkmark & -- & 1,156 \\
Time-based & \checkmark & -- & \checkmark & 1,155 \\
Time-based & \checkmark & -- & -- & 1,072 \\
\bottomrule
\end{tabular}

\label{reward_exploration}
\end{subtable}
\end{table}
$ $

\noindent \textit{GCN Update Function:} Here we test the effects of several design decisions for the StagedGCN. Namely, we test the utility of skip connections, layer normalization and whether it is best to take the graph embedding as the formula's root node embedding only or as the average over the all nodes within the graph.  As shown in Table~\ref{tab:ablation}, skip connections and layer normalization are key to performance. This is likely due to the size and depth of logical formulas (e.g., in Mizar there are logical formulas with over 2000 nodes in their graph representation). 




\subsection{Bounded Rewards and Exploration Effect}

As mentioned in Section \ref{reward_info}, there are many different ways to structure the reward function used to train \trail. In this experiment, we study the effect on performance that alterations to the reward function have. In particular, using the MPTP dataset we examine the differences when the reward is (i) unnormalized, (ii) normalized by the inverse of the time spent by a traditional reasoner, and (iii) normalized as in (ii) but is restricted to fall within the range $[1, 2]$. For normalized vs unnormalized rewards, we also study the effect of using a step-based (number of steps used to find a proof) compared to using a time-based reward (total time taken to find a proof).

Table~\ref{reward_exploration} shows the performance of TRAIL under different settings when starting from the same randomly initialized model. The results clearly show that the time-based reward is more effective (perhaps due to it more holistically capturing the complexity of finding a proof). It also shows that normalization is very helpful for solving more problems. Because time-based rewards are more effective than step-based rewards, we use the time-based normalized reward to demonstrate the effect of reward bounding and limiting exploration. When limiting exploration, the temperature parameter is set to 1, i.e., $\tau = 1$, which has the effect that \trail always takes the prediction of the policy network when selecting an action. Without bounding the reward, \trail solves 58 less problems, and when limiting its exploration ability \trail solves 57 less problems. When disabling both bounded rewards and exploration, \trail solves 141 less problems on MPTP benchmark.


\begin{table*}
\caption{Example problems from MPTP benchmark with TRAIL vs. E-prover runtime (seconds) and proof steps. Timeout is used to indicate that no proof could be found}
\centering
\begin{tabular}{ll|cc|cc}
\toprule
                          &                                                                  & \multicolumn{2}{c|}{TRAIL} & \multicolumn{2}{c}{E-Prover} \\
\midrule
Problem File                  &     Theorem                                                             & Time        & Steps       & Time           & Steps       \\
\midrule
relat\_1\_\_t147          & Relations and Their Basic Properties                             & 0.8         &    8         & 0.01           &       13      \\
tops\_1\_\_t31 & Subsets of Topological Spaces                             & 1.2         &    12         & 0.02           &     18        \\
yellow\_6\_\_t20          & Moore-Smith Convergence                                          & 6.2         &   61          & 0.03           &     37        \\
\midrule
connsp\_2\_\_t31          & Locally Connected Spaces                                         & 3.9         & 37          & 996.1          & 70          \\
ordinal1\_\_t37           & The Ordinal Numbers                                              & 6.5         &    62         & 1455.6         &       42      \\
orders\_2\_\_t62          & Kuratowski -- Zorn Lemma                                         & 41.3       &    396         & 1430.8        &  109           \\
\midrule
waybel\_0\_\_t14          & Lattices  Directed Sets, Nets, Ideals, Filters, and Maps         & 15.3        &     140        & Timeout        &      N/A       \\
wellord1\_\_t23 & The Well Ordering Relations                                      & 16.5        &    147         & Timeout        &       N/A      \\
tex\_2\_\_t41             & Maximal Discrete Subspaces of Almost Discrete Topological Spaces & 61.7        &  577           & Timeout        &  N/A \\
\midrule
tmap\_1\_\_t6 &  Continuity of Mappings over the Union  of Subspaces & Timeout        &     N/A        & 9.4        &   76 \\
xboole\_1\_\_t83&    Boolean Properties of Sets    & Timeout        &     N/A        & 23.0        &   56 \\
zfmisc\_1\_\_t99&   Some Basic Properties of Sets  & Timeout        &     N/A        & 31.2        &   53 \\

\midrule
mcart\_1\_\_t51& Tuples, Projections and Cartesian Products  & Timeout        &  N/A           & Timeout        &  N/A \\
compts\_1\_\_t29& Compact Spaces  & Timeout        &  N/A           & Timeout        &  N/A \\
\bottomrule
\end{tabular}
\label{tab:qualitative}
\end{table*}

\subsection{Qualitative Analysis}
Table \ref{tab:qualitative} shows example problems from MPTP benchmark along with the runtime and number of steps taken to find a proof by \trail in comparison to E (in auto mode). In this experiment, we ran E for 30 minutes to lessen the potential for timing out. On easy problems with shorter proofs such as \textit{relat\_1\_\_t147} and \textit{tops\_1\_\_t31}, E is much faster in terms of time as compared to \trail (0.01 and 0.02 seconds versus 0.8 and 1.2 seconds for \trail) even though \trail's proofs are sometimes shorter; 8 versus 13 on \textit{relat\_1\_\_t147}. As problems get harder (such as with \textit{orders\_2\_\_t62} which requires more than a hundred steps), \trail was able to find a proof much faster compared to E (41.3 versus 1430 seconds) at the expense of having a longer proof. As Table \ref{tab:qualitative} shows, this pattern continues where E could not find a proof within 30 minutes where \trail solved these problmes in less than 100 seconds. The last two categories are for problems that \trail does not solve, and problems that neither E nor \trail could solve, which is a category we plan to investigate in  future work.

\section{Related Work}

Several approaches have focused on the sub-problem of premise selection (i.e., finding the axioms relevant to proving the considered problem) \cite{AlamaHKTU-jar14-premises-corpus-kernel,Blanchette-jar16-premises-isabelle-hol,Alemi-NIPS16-deepmath,WangTWD-NIPS17-deepgraph}. 
As is often the case with automated theorem proving, most early approaches were based on manual heuristics \cite{hoder2011sine,roederer2009divvy} and traditional machine learning \cite{AlamaHKTU-jar14-premises-corpus-kernel}; though some recent works have used neural models \cite{Alemi-NIPS16-deepmath,WangTWD-NIPS17-deepgraph,rawsondirected2020,crouse2019improving,DBLP_conf_lpar_PiotrowskiU20}. 
Additional 
research has used learning to support interactive theorem proving \cite{Blanchette-jar16-premises-isabelle-hol,Bancerek-JAR2018-mml}.

Some early research applied (deep) RL 
for guiding inference \cite{TaylorMSW-FLAIRS07-cyc-rl}, planning, and machine learning techniques for inference in relational domains
\cite{surveyRLRD}. 
Several papers have considered propositional logic or other decidable FOL fragments, which are much less expressive compared to \trail. 
Closer to \trail are the approaches described in
\cite{KalUMO-NeurIPS18-atp-rl,zombori2020prolog} where RL is combined with Monte-Carlo tree search
for theorem proving in FOL. However they have some limitations: 
1)~Their approaches are specific to tableau-based reasoners and thus not suitable for theories with many equality axioms, which are better handled in the superposition calculus \cite{bachmair1994refutational}, and 
2)~They rely upon simple linear learners and gradient boosting 
as policy and value predictors. 

Our work also aligns well with the recent proposal of an API for deep-RL-based interactive theorem proving in HOL Light, using imitation learning from human proofs \cite{BaLoRSWi-CoRR19-holist}. 
Their work describes a learning environment for HOL Light with some neural baselines for  higher-order theorem proving. Unlike this paper, \trail targets first-order logic, built to guide state-of-the-art theorem provers such as E and Beagle, and have an efficient graph neural network based vectorization technique for logic formulas.    

Non-RL-based approaches using deep-learning to guide proof search include \cite{ChvaJaSU-CoRR19-enigma-ng,LoosISK-LPAR17-atp-nn,paliwal2019graph}. These approaches differ from ours in that they seed the training of their networks with proofs from an existing reasoner. In addition, they use neural networks during proof guidance to score and select available clauses with respect \emph{only} to the conjecture. Recent works have focused on addressing these two strategies. For instance, \cite{piotrowski2019guiding} explored incorporating more than just the conjecture when selecting which inference to make with an RNN-based encoding scheme for embedding entire proof branches in a tableau-based reasoner. However, it is unclear how to extend this method to saturation-based theorem provers, where a proof state may include thousands of irrelevant clauses. Additionally, \cite{aygun2020learning} investigated whether synthetic theorems could be used to bootstrap a neural reasoner without relying on existing proofs. Though their evaluation showed promising results, it was limited to a subset of the TPTP \cite{tptp} that excluded equality. It is well known that the equality predicate requires much more elaborate inference systems than resolution \cite{bachmair1998equational}, thus it is uncertain as to whether their approach would be extensible to full equational reasoning.

Approaches for first-order logic also differ in tasks they were evaluated on, with some evaluated on offline tasks such as premise selection \cite{WangTWD-NIPS17-deepgraph,crouse2019improving,rawsondirected2020}, length prediction \cite{Xavier_Glorot2019}, and a few in online proof guidance \cite{olvsak2019property,rawson2019neurally,rawsonreinforced,rawson2021automated}. In online proof guidance, which our work targets, existing work are based on simpler tableaux based reasoners \cite{olvsak2019property,rawson2019neurally,rawson2021automated}. Unlike these approaches,  \trail targets guiding efficient, more capable saturation-based theorem provers. Furthermore, \trail leveraging the unique characteristics of DAG-shaped logic formulas and taking into account the time-limited nature of theorem proving, \trail was able to achieve state-of-the-art performance on two theorem proving benchmarks. 
A fundamental problem in using neural approaches to proof guidance is learning a suitable representation of logical formulas. Representations learned with graph neural networks have become the method of choice recently due to their faithfulness in preserving syntactic and semantic properties inherent in logical formulas. 
However, approaches that target specific logics such as propositional logic \cite{selsam2018learning,Xavier_Glorot2019} and fragments of first-order logic \cite{barcelo2019logical}, fail to preserve properties specific to first-order logic. Recent work tried to address this limitation by  preserving properties such as invariance to predicate and function argument order \cite{olvsak2019property,rawsondirected2020} and variable quantification \cite{paliwal2019graph,olvsak2019property,rawson2019neurally,rawsondirected2020,crouse2019improving,rawson2021automated}. Other approaches specifically target higher-order logics, which have different corresponding graph structures from first-order logic \cite{WangTWD-NIPS17-deepgraph,paliwal2019graph}.


\section{Conclusions}
\label{sec:conc}

In this work, we introduced \trail, an end-to-end neural approach that uses deep reinforcement learning to learn effective proof guidance strategies from scratch. \trail leverages (a) a novel graph neural network that takes into account the time-limited nature of theorem proving and the unique characteristics of logic formulas, (b)  a novel representation of the state of a saturation-based theorem prover, and (c) an attention-based proof guidance policy. Through an extensive experimental evaluation, we have shown that \trail outperforms all prior reinforcement learning-based approaches on two standard benchmark datasets. Furthermore, to the best of our knowledge, \trail is the first system to learn proof guidance strategies from scratch and outperform a state-of-the-art traditional theorem prover on a standard theorem proving benchmark. For future work, we plan to extend \trail to more expressive logic formalisms such as higher-order logic. 

\ifCLASSOPTIONcaptionsoff
  \newpage
\fi



\bibliographystyle{IEEEtran}
\bibliography{references}

\end{document}